\pdfoutput=1

\documentclass[11pt]{article}

\usepackage[preprint]{acl}

\usepackage{amsmath}
\usepackage{amssymb}
\usepackage{multicol}
\usepackage{multirow}
\usepackage{pythonhighlight}
\usepackage{booktabs}
\usepackage{url}
\usepackage{times}
\usepackage{latexsym}
\usepackage{enumitem}

\usepackage[T1]{fontenc}

\usepackage[utf8]{inputenc}

\usepackage{microtype}

\usepackage{inconsolata}

\usepackage{graphicx}

\iftrue

\fi
\title{QUBE: Enhancing Automatic Heuristic Design via Quality-Uncertainty Balanced Evolution
}

\author{\vspace{-3mm} 
\centerline{Zijie Chen$^{1,2}$ ~~ Zhanchao Zhou$^{1,2}$ ~~ Yu Lu$^{2}$ ~~ Renjun Xu$^1$ ~~ Lili Pan$^{\dagger,3}$ ~~ Zhenzhong Lan$^{\dagger,2}$}
\\\\ 
\centerline{\normalfont{$^1$Zhejiang University} \quad {$^2$Westlake University}} \\ 
\centerline{{$^3$University of Electronic Science and Technology of China}}
\\ 
\small \centerline{\texttt{\{chenzijie, lanzhenzhong\}@westlake.edu.cn , lilipan@uestc.edu.cn}} }

\begin{document}
\maketitle
\begin{abstract}
Solving NP-hard problems traditionally relies on heuristics, yet manually designing effective heuristics for complex problems remains a significant challenge. While recent advancements like FunSearch have shown that large language models (LLMs) can be integrated into evolutionary algorithms (EAs) for heuristic design, their potential is hindered by limitations in balancing exploitation and exploration. We introduce Quality-Uncertainty Balanced Evolution (QUBE), a novel approach that enhances LLM+EA methods by redefining the priority criterion within the FunSearch framework. QUBE employs the Quality-Uncertainty Trade-off Criterion (QUTC), based on our proposed Uncertainty-Inclusive Quality metric, to evaluate and guide the evolutionary process. Through extensive experiments on challenging NP-complete problems, QUBE demonstrates significant performance improvements over FunSearch and baseline methods. 
Our code are available at \url{https://github.com/zzjchen/QUBE\_code.}
\end{abstract}
\renewcommand{\thefootnote}{$^\dagger$}
\footnotetext{Corresponding author.}

\section{Introduction}
\label{introduction}

Many mathematical science problems are NP-complete, making them extremely challenging to solve but easy to evaluate~\cite{romera2024mathematical}. Evolutionary Algorithms (EAs) are widely used to optimize heuristics for such problems~\cite{liu2023algorithm, Mei2023Explainable}. Recently, large language models (LLMs) have demonstrated remarkable capabilities in code generation \cite{austin2021program,chen2021evaluating,li2023starcoder}, opening up new avenues for hyper-heuristic algorithms. These methods, termed ``LLM+EA" methods, leverage LLMs as variation operators within EAs, achieving promising results across diverse domains~\cite{chen2024evoprompting,zheng2023can,nasir2024llmatic,wang2024efficient}. A notable example is FunSearch \cite{romera2024mathematical}, which discovers high-quality heuristics through approximately 2.5 million evolutionary steps in a multi-population EA framework.

\begin{figure}[t]
    \centering
    \includegraphics[width=0.98\linewidth]{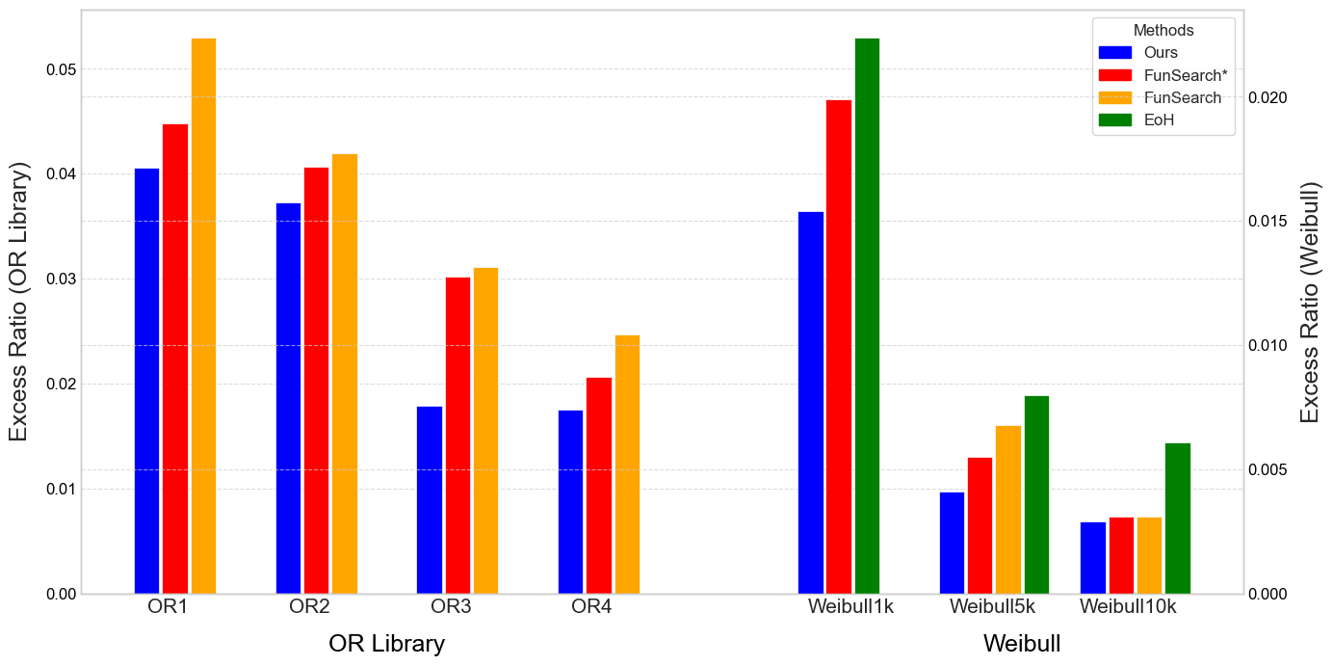}
    \caption{ Experiment results of our method on online bin packing, the performance is evaluated with ``Excess Ratio" and the lower the better. Our method can steadily find better heuristics than all baselines.
    }
    \label{fig: figure1}
\end{figure}

Theoretically, to optimize heuristics in a ``function space", a method should excel in two key aspects: exploitation (deepening search in promising regions) and exploration (broadening search in unknown regions). However, achieving this balance long remains an open challenge~\cite{weng2020exploration}. Through analysis, we observe that the priority criterion behind FunSearch's evolution process hinders it from exploiting upon current status and performing useful exploration within the function space, resulting its struggle.

To address these issues, we propose Quality-Uncertainty Balanced Evolution (QUBE), a novel approach that enhances the heuristic evolution in FunSearch by redefining the priority criterion of the evolutionary process. Central to QUBE is the Quality-Uncertainty Trade-off Criterion (QUTC), which is based on our proposed Uncertainty-Inclusive Quality (UIQ) metric, with inspiration drawn from the Upper Confidence Bound \cite{10.1016/0196-8858(85)90002-8,auer2002using}. QUBE is experimented on both standard combinatorial optimization problems and complex challenges like the cap set problem. As shown in Figure \ref{fig: figure1}, it shows significant superiority over baseline methods\footnote{We only show results for online bin packing here, please refer to Appendix \ref{appn: fig1 more}for more results. See Section \ref{Experiments} for experimental details.}.

We summarize our contributions as follows:
\begin{enumerate}[leftmargin=2em, itemindent=0em, itemsep=0pt, parsep=0pt, topsep=0pt, partopsep=0pt]
    \item We identify that FunSearch's priority criterion limits its search performance, stemming from an insufficient balance between exploitation and exploration in heuristic evolution.
    \item We propose QUBE, an LLM+EA method that employs our propriety criterion QUTC to automatically balance exploitation and exploration throughout the evolutionary process.
    \item Experimental results across multiple NP-complete problems demonstrate significant improvements: reduction in excess bin usage for online bin packing (OBP), enhanced solution quality for traveling salesman problem (TSP), and larger cap set discoveries.
\end{enumerate}

\section{Related Work} 
\subsection{Heuristics for Math Problems}
Heuristics are typically used to search solutions for NP-hard problems such as the Traveling Salesman Problem (TSP)~\cite{liu2023algorithm}, online bin packing (OBP)~\cite{coffman1984approximation}, cap set problem~\cite{grochow2019new,tao2006additive} etc. They guide the search direction to find relatively good solutions within a reasonable time. While it's hard to hand craft a good heuristic, hyper-heuristics algorithms~\cite{Burke2003hyper} like EA can automatically optimize heuristics from a trivial on \cite{Jia2023Learning, Mei2023Explainable}. Since the boost of deep learning, various relevant methods have been used to assist EA \cite{Bengio2021machine,Hudson2022Graph,Hottung2020heuristic}.
\subsection{LLM+EA}
The effectiveness of EA heavily relies on the ability of variation operators to generate diverse and promising new candidates, a process that typically demands substantial domain-specific knowledge \cite{Michael2010open}. Recent research has explored the integration of EAs with LLM's generative potential, termed LLM+EA methods \cite{Lehman2024}. These methods leverage the few-shot generation capabilities of LLMs as variation operators, extending their applications to diverse domains such as neural architecture search \cite{chen2024evoprompting}, text-based tasks \cite{meyerson2023language}, optimization \cite{brahmachary2024large}, and molecular design \cite{wang2024efficient}.

Subsequent studies have focused on refining LLM+EA methodologies by enhancing prompting and generation strategies. For instance, EoH \cite{liu2024evolution} introduces five distinct prompts tailored for exploration and modification, moving beyond the single fixed prompt used in earlier approaches. Additionally, EoH suggests that LLMs should first generate a textual description before implementing code. Similarly, ReEvo \cite{ye2024reevo} incorporates LLM reflection into the process, enabling the model to generate improved samples based on insights derived from historical data. Despite these advancements, existing LLM+EA methods still face challenges in scalability, efficiency, and their applicability to more complex problems.

\subsection{FunSearch and Beyond}

Existing LLM+EA methods have predominantly operated on a limited scale, typically generating fewer than 10,000 samples throughout the evolutionary process. These approaches have not yet fully leveraged the generative potential of LLMs or the evolution power of EAs. As a result, their applications have largely been confined to conventional combinatorial optimization problems, such as the TSP and OBP, which require relatively few evolutionary steps to yield meaningful results.

In contrast, FunSearch \cite{romera2024mathematical} represents a significant leap in scaling LLM+EA methods, generating approximately 2.5 million samples during its evolutionary process. FunSearch extends beyond theoretical and mathematical domains, addressing complex and significant challenges such as the cap set and admissible set problems. By significantly scaling up the generation of sample, FunSearch has demonstrated that LLM+EA algorithms can achieve state-of-the-art (SOTA) solutions to exceptionally difficult problems, surpassing the capabilities of all prior LLM+EA methods.

\label{FunSearch}
\begin{figure*}
    \centering
    \includegraphics[width=0.98\textwidth]{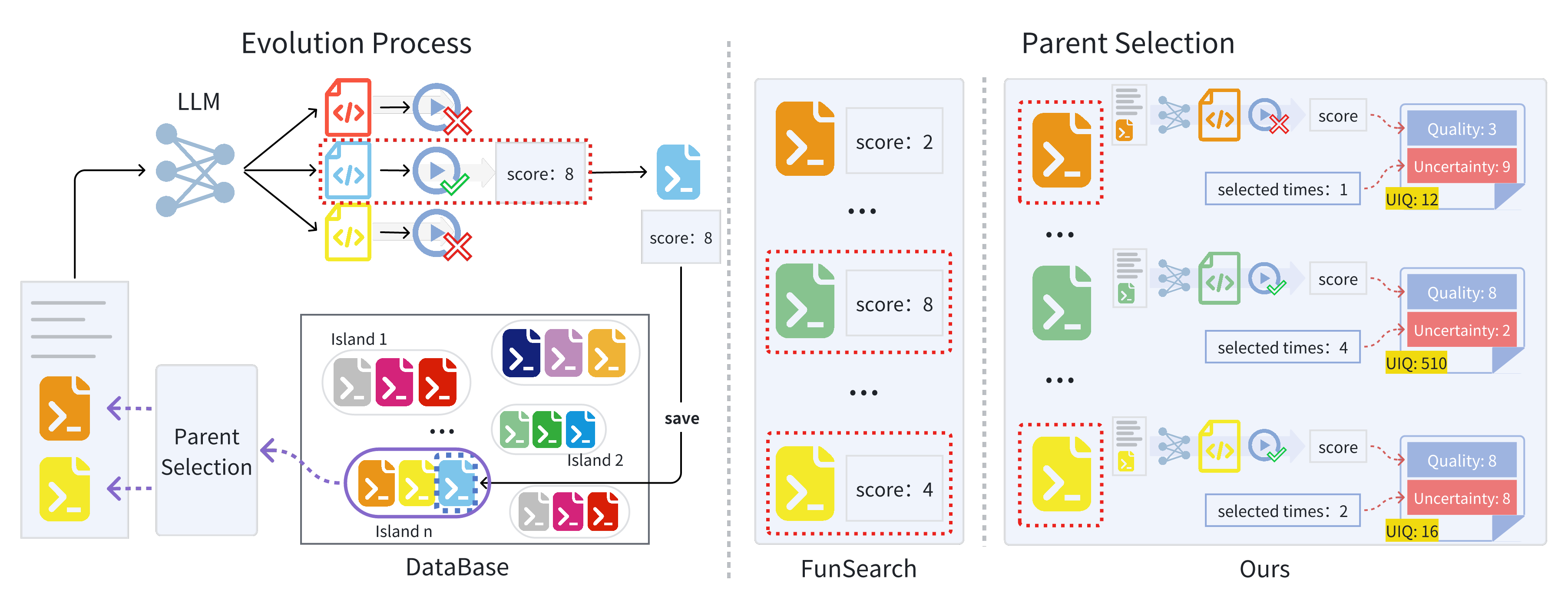}
    \caption{Illustration of QUBE. We manipulate the parent selection procedure in FunSearch's evolution process. \textbf{Left}:  The overall evolution process of our method and FunSearch. \textbf{Right}: At each timestep, FunSearch selects parents based on the score of each sample. Our method selects parents based on our quality measure, UIQ. The uncertainty of a sample's quality is acquired from the number of times it is selected as parents.
    }
    \label{fig: pipeline}
\end{figure*}

\section{Thoroughout Examining Exploration and Exploitation in FunSearch}
In this section, 
we first provide an overview of FunSearch and elaborate on two important details: parent selection during each evolution step, and island reset that periodically takes place.  A priority criterion affects these core details is identified. Then, we define exploitation and exploration in FunSearch and analyze how the priority criterion affects the balance between exploitation and exploration. Finally, we empirically show FunSearch's deficiency in both exploitation and exploration.

\subsection{Overview of FunSearch}
FunSearch is an LLM+EA method designed to evolve heuristics of some problems, represented as Python functions.  It employs a frozen LLM as a variation operator within an EA framework that utilizes multiple populations, or "islands."
 An overview of FunSearch's evolutionary process is illustrated in the left part of Figure \ref{fig: pipeline}.

At each step, a randomly selected island undergoes the evolution process. Two parent samples are chosen from this island, and the LLM is prompted to generate new samples using the parents as few-shot examples. These newly generated samples are evaluated for performance, and only those that execute without Python exceptions or timeouts are retained on the island. Periodically, FunSearch resets underperforming islands by deleting all their samples and reinitializing them with the best-performing sample from a high-performing island. Specifically, half of the islands with the lowest performance are reset in this manner.

Central to FunSearch is a priority criterion that determines the selection of parent samples and identifies islands requiring a reset. FunSearch defines this priority criterion as the samples' scores. Specifically, the probability of a sample being selected as a parent is proportional to the exponential of its score.  Similarly, an island is reset at each island reset interval if its highest-scored sample underperforms at least half of the other islands. The priority criterion of FunSearch ensures that the evolutionary process prioritizes high-performing samples while maintaining some diversity across populations.

\subsection{Exploration and Exploitation in FunSearch}
The primary objective of FunSearch is to identify high-performance heuristics through iterative sampling. To achieve this, the method must effectively exploit the known function space by continuously generating new samples with improved performance. However, restricting the search to a limited region of the function space makes it challenging to discover highly effective heuristics. Therefore, in addition to exploiting well-known regions, the method must also explore less-explored areas, even if they initially appear unpromising, by generating diverse samples. These two complementary strategies are referred to as \textit{exploitation} and \textit{exploration}, respectively.

We show how the priority criterion influences the balance between exploitation and exploration, which in turn affects the overall performance of LLM+EA methods such as FunSearch. At each evolutionary step, the priority criterion is used to select two parent samples to guide a frozen LLM sampler in generating new samples. To maximize exploitation, the criterion should prioritize parents likely to produce high-performance offspring. Conversely, to encourage exploration, it should also consider parents with uncertain outcomes, enabling the discovery of novel regions in the function space.

For methods that incorporate island reset mechanisms, such as FunSearch, the priority criterion also plays a critical role in determining which islands to reset. Islands that have extensively explored their regions but consistently produce heuristics with relatively low scores should be reset to prioritize exploitation. Conversely, islands with low performance but incomplete exploration should be preserved to encourage further exploration. 

Ultimately, the priority criterion must strike a careful balance between exploitation and exploration, as overemphasis on either strategy can compromise the effectiveness of the other.

\subsection{Quantitative Assessment of Exploration and Exploitation}
\label{analysis}

\begin{figure}[!t]
    \centering
    \includegraphics[width=0.98\columnwidth]{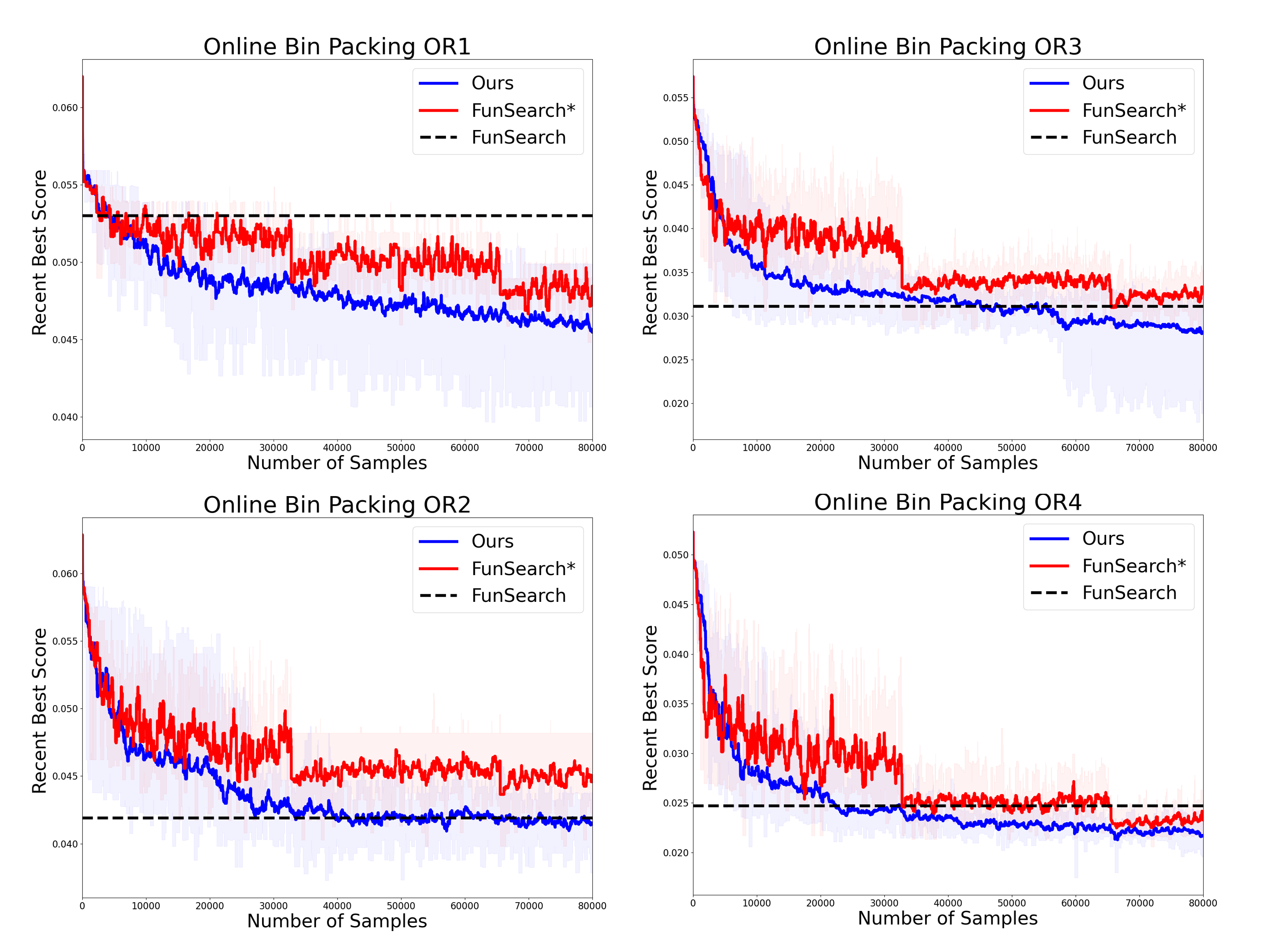}
    \caption{The ``Recent Best Score" of FunSearch exhibits plateaus in later stages, indicating challenges in effectively exploiting known regions. In contrast, our method consistently generates higher-scoring samples, demonstrating superior exploitation capabilities.}
    \label{fig: recentbest}
\end{figure}

To quantify exploitation and exploration, we introduced two evaluation metrics: "Recent Best Score" and "Recent Proportion of Change". In practice, we set K=500 for both metrics.

\noindent\textbf{Recent Best Score}: It measures the highest score among the $K$ most recently generated samples, reflecting the method's ability to exploit known regions effectively. A higher "Recent Best Score" indicates successful exploitation of high-performing regions in the function space.

\noindent\textbf{Recent Proportion of Change}: It computes the average ``proportion of change" observed in correct programs across the most recent K samples. The ``proportion of change" is quantified as the token-level edit distance between a generated sample and its nearest parent, normalized by the length of the sample. This metric indicates exploration, as  a higher "Recent Proportion of Change" indicates the generation of novel and diverse samples, suggesting the discovery of previously unexplored regions in the function space.

\begin{figure}[!t]
    \centering
    \includegraphics[width=0.98\columnwidth]{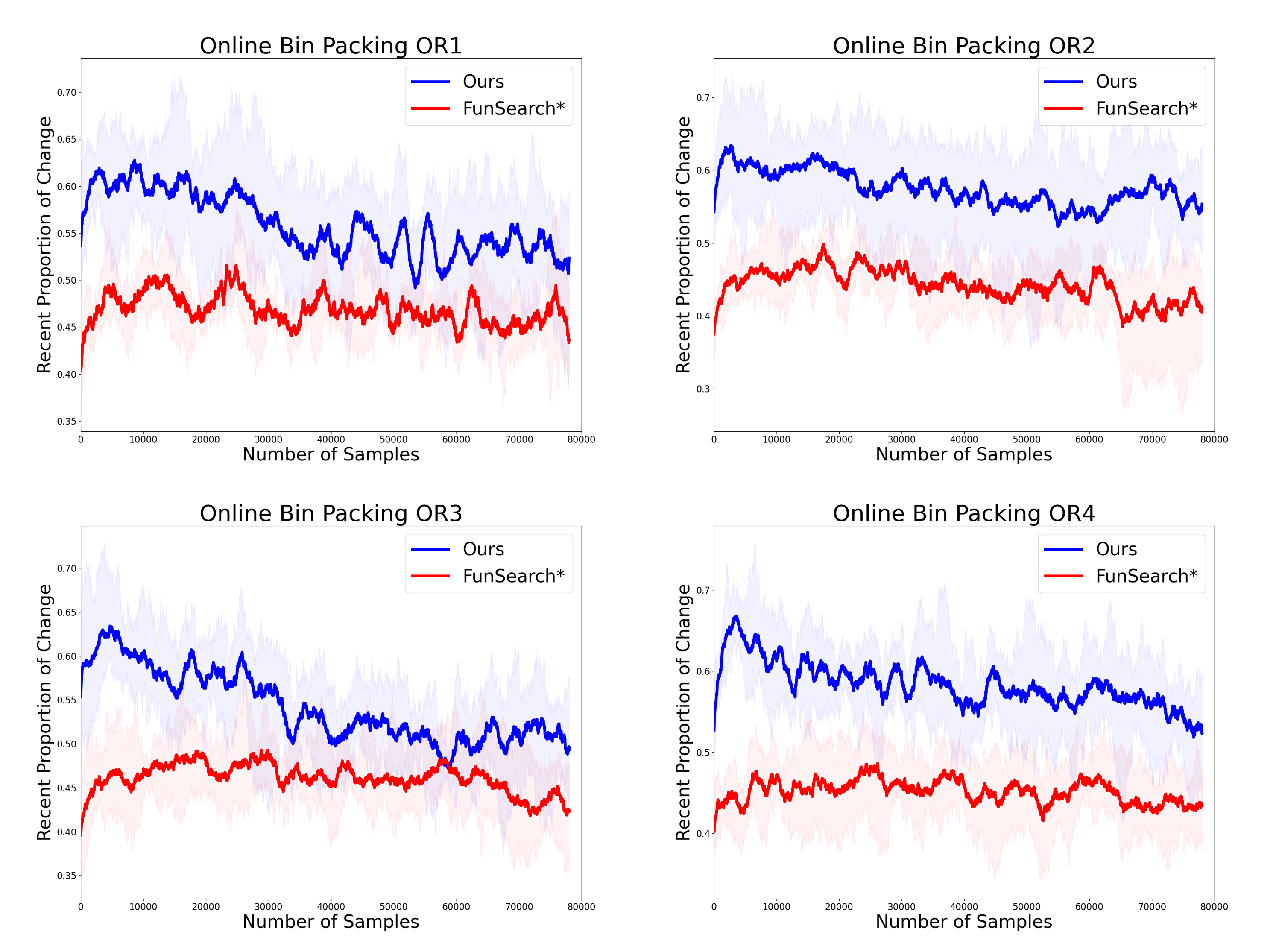}
    \caption{FunSearch has a consistently low ``Recent Proportion of Change", reflecting its limited overall exploration of the function space. In contrast, our method demonstrates both a broader scope and a more intelligent exploration strategy, enabling more effective discovery of promising regions.}
    \label{fig: recentpoc}
\end{figure}

In Figure \ref{fig: recentbest}, the "Recent Best Score" of FunSearch is visualized in red. The presence of plateaus in the curve indicates slow improvements during the later stages of evolution, suggesting that it struggles to effectively exploit known regions. This limitation arises because it uses score prioritization as the priority criterion during evolution, which does not necessarily correlate with the performance of newly generated samples.

Furthermore, despite employing techniques such as multi-population evolution, FunSearch's exploration strategy is indiscriminate, as it randomly explores the function space without considering whether the current region is promising. This is evident in Figure \ref{fig: recentpoc}, where FunSearch's exploration remains constant and relatively low throughout the evolutionary process. A more intelligent exploration strategy should prioritize regions with a higher likelihood of containing high-scoring samples while reducing exploration in less promising areas. Such a strategy would naturally emphasize exploration in the early stages when most regions remain unexplored, and gradually shift focus toward exploitation as fewer promising regions are left undiscovered. This adaptive approach would allow for a better balance, ultimately increasing the likelihood of generating higher-scoring samples.



\section{Quality-Uncertainty Balanced Evolution of Heuristics}
To balance exploration and exploitation in hubristic evolution, we propose Quality-Uncertainty Balanced Evolution (QUBE). In the following, we first outline the overall framework of our method. Next, we introduce our priority criterion Quality-Uncertainty Trade-off Criterion (QUTC), which is based on our proposed Uncertainty-Inclusive Quality (UIQ) for evaluating samples. Finally, we demonstrate how QUBE integrates QUTC into key components of the evolutionary process, including parent selection and the island reset procedure.

\subsection{Overall Framework}
At a macro level, the overall structure of our method (Figure \ref{fig: pipeline}'s left) aligns with that of FunSearch. Both approaches aim to evolve a Python function that serves as a heuristic within a search algorithm. The performance of each function sample $c$ is evaluated deterministically by executing the search algorithm on a predefined set of test instances, yielding a score $s(c)$. All samples are stored in a database $\mathbb{D}$, which consists of $n\geq 1$ islands. Each island $\mathbb{I}$ maintains an independent population for evolution, with no communication between islands except during island resets. Furthermore, each island is organized into multiple clusters. Within a cluster $\mathbb{C}$, program samples that yield identical results on all test instances are grouped together. Consequently, all samples within a cluster share the same score, denoted as $s(\mathbb{C})$, repurposing the function notation for clarity.

At each evolutionary step, our method randomly selects an island $\mathbb{I}$ to generate new samples uniformly. Two parent samples are chosen from $\mathbb{I}$ using our priority criterion, QUTC. These parent samples are then provided as few-shot examples to the LLM, which generates new samples. After evaluation, the newly generated samples are stored back into the same island $\mathbb{I}$. Periodically, after every $T_{reset}$ sample generation, our method identifies and resets half of the underperforming islands with same procedure as FunSearch. This reset mechanism ensures a balance between exploration and exploitation by revitalizing underperforming regions of the search space.

\subsection{Quality-Uncertainty Trade-off Criterion}
To effectively balance exploitation and exploration, our priority criterion QUTC must identify samples that offer evolutionary advantages, specifically those likely to produce high scores in newly generated samples, while also considering less-explored regions of the search space, represented by samples that have been visited less frequently. In practice, we observed significant similarity among samples within the same cluster. Thus, QUTC prioritizes clusters as a whole rather than individual samples, ensuring a more efficient and scalable approach to guiding the evolutionary process.

We first introduce UIQ, the metric we used to assess the quality of samples within a cluster. At each timestep $t$, we compute for each cluster $\mathbb{C}$ the mean score of all offspring generated using samples from $\mathbb{C}$ as parents. This is formally expressed as: 
\begin{equation}\label{eq: q}
    Q_t(\mathbb{C})=\frac{1}{\sum\limits_{c\in\mathbb{C}}|\mathbb{P}_{c,t}|}\sum\limits_{c\in \mathbb{C}}{\sum\limits_{a\in \mathbb{P}_{c,t}}{s(a)}}
\end{equation}
where $\mathbb{P}_{c,t}$ is a collection of all samples generated with $c$ as a parent 
before timestep $t$. $Q_t(\mathbb{C})$ estimates the expected performance of offspring produced by samples in $\mathbb{C}$, enabling the identification of clusters that exhibit evolutionary advantages.

Inspired by Upper Confidence Bound (UCB), we incorporate uncertainty into $Q_t(\mathbb{C})$, resulting in UIQ. 
Let $N_t(\mathbb{C})$ be the number of times samples in cluster $\mathbb{C}$ are used as parents before timestep $t$. We define UIQ as:
\begin{equation}\label{eq: uiq}
    \Tilde{Q}_t(\mathbb{C})=Q_t(\mathbb{C})+k\sqrt{\frac{\mathrm{ln}\ t}{N_t(\mathbb{C})}}
\end{equation}
where $k$ is a hyperparameter. 

As evident from its formulation, UIQ combines an estimate of a cluster's evolutionary quality with the uncertainty of that estimate. Thus QUTC can automatically balance the exploitation of high-performing regions and the exploration of less-explored promising areas in the search space by prioritizing clusters with higher UIQ values.

\subsection{Quality-Uncertainty Balanced Evolution}
Our method QUBE incorporates QUTC into the parent selection at each evolution step and the evaluation of islands at each island reset.

As illustrated in the right part of Figure \ref{fig: pipeline}. After an island $\mathbb{I}$ is selected to evolve new samples at each timestep $t$, we identify 2 clusters in $\mathbb{I}$ with the highest UIQ according to QUTC. We select one sample per cluster to serve as parents for this step. Specifically, let $l_c$ be the length of sample $c$ measured by the number of characters, and $\Tilde{l}_c=\frac{\mathrm{max}_{a\in\mathbb{C}}{\{l_a\}}-l_c}{\mathrm{min}_{a\in\mathbb{C}}{\{l_a\}}+1e^{-6}}$. The probability of chosen $c$ within a cluster is proportionate to $\mathrm{exp}(\frac{\Tilde{l}_c}{T_{prog}})$, where $T_{prog}>0$ is a hyperparameter. 

At each island reset interval, we evaluate the quality of each island using the cluster with the highest UIQ within that island. Islands whose highest UIQ falls below the median among all islands are selected for reset. For each reset island, reinitialization is performed by selecting a random sample from the best cluster of a randomly chosen remaining island, ensuring a promising restart for further evolution.

\section{Experiments}
\label{Experiments}

\subsection{Implementation Details}
\label{implement detail}

We implement an asynchronous system on a single server with 8 NVIDIA A100 GPUs and 2 Intel(R) Xeon(R) Platinum 8358 CPUs. On each GPU, an LLM inference service is set up locally using the SGLang \cite{zheng2024sglang} framework. This segregates LLM inference from the entire system, maximizing the advantages of asynchronous concurrency. We use OpenCoder-8B-Instruct \cite{huang2024opencoder} throughout our experiment, while also experiment with Deepseek-coder \cite{Guo2024deepseekcoder} to ablate the influence of LLM. We provide our prompt for LLM in Appendix \ref{appn: prompt}.

The remaining components of our implementation operate in parallel through multiprocessing. The database is shared and accessible to all processes. Our samplers iteratively retrieve parent samples (examples) from the database and submit requests to the backend LLM services. 
Upon the generation of new samples, evaluators are called by the samplers to assess these samples before their storage in the database.
Other hyperparameter settings are shown in Table \ref{tab: fun_detail} in Appendix. Note for TSP, a very small amount of sample is required to get relatively good result. Thus we use only 1 island and removed the island reset for TSP.

\subsection{Experiment problems}
\label{experiment problems}
We assessed the performance of our method on three NP-complete problems:

\noindent\textbf{Online Bin Packing}: 
We focus on its online scenario, where each item is packed as it arrives. We conduct experiments on the OR-Library \cite{beasley1990or}, which comprises four datasets of online bin packing instances (OR1 to OR4). We also tested our method on generated instances from Weibull distribution. Identical to FunSearch \cite{romera2024mathematical}, our method evolves the heuristics within a local-search algorithm. 
We evaluate the methods using the fraction of excess bins used over the L2 lower bound \cite{martello1990lower} of the optimal offline bin packing solution, a metric we refer to as the ``excess ratio". 

\noindent\textbf{Cap Set}: 
The cap set problem finds the largest ``cap set", which is a set of vectors in $\mathbb{Z}_3^n$ such that the sum of any three vectors is not zero. 
    As with FunSearch \cite{romera2024mathematical}, our method evolves a priority function that assigns a rank to each vector in  $\mathbb{Z}_3^n$, which guides a greedy construction of cap sets.  We carry out experiments for $n=8$, and use the size of the largest cap sets found as performance.
    
 \noindent\textbf{Traveling Salesman Problem}: TSP is a combinatorial optimization problem, which finds shortest routes that visit all given locations once and return to the starting point. We experimented with our method on 3 settings, namely TSP20, TSP50 and TSP100, following previous works \cite{kool2018attention,liu2024evolution}. Identical to \cite{liu2024evolution}, our method is used to evolve the objective function in the perturbation stage of a guided local search algorithm \cite{voudouris2010guided}. The relative distance between the acquired solution and the optimal solution calculated by Concorde \footnote{https://www.math.uwaterloo.ca/tsp/concorde.html} is used to assess the performance of each method, which we also termed as ``excess ratio".

Each experiment is run 10 times, and the best result among all is reported unless otherwise specified. In the ablation study, we include the average performance as well as the standard deviation to examine if the results are robust. Please refer to Appendix \ref{appn: more exp detail} for more details on how the data for each problem are generated. The code specification of each task is available at Appendix \ref{appn: specification}.

\subsection{Baselines}
\label{exp baseline}
We compared our method with extensive baselines, including:
(1) \noindent\textbf{FunSearch}: For comparison, we use directly the performance on online bin packing and cap set reported in FunSearch ~\cite{romera2024mathematical}. Since we are not using the same LLM and hardware compared with FunSearch~\cite{romera2024mathematical}, we reproduced the FunSearch method on our GPU server according to our implementation details, denoted as \textbf{FunSearch*}. 
(2) \textbf{EoH}: 
For online Bin Packing and TSP, we also compared the result of our method with the result of EoH \cite{liu2024evolution,zhang2024understanding}.

\begin{table*}[t]
    \centering
    \resizebox{1\textwidth}{!}{
    \begin{tabular}{|l|c c c c c c c| c| c c c|}
        \hline
        \multirow{2}{1em}{} & \multicolumn{7}{|c|}{Online Bin Packing ($\downarrow$)}& Cap Set ($\uparrow$)& \multicolumn{3}{|c|}{TSP ($\downarrow$)} \\
        & OR1 & OR2 & OR3 & OR4 & Weibull 1k& Weibull 5k & Weibull 10k  &n=8 & TSP20& TSP50& TSP100\\
        \hline
        \hline
        Ours & \textbf{4.06\%}& \textbf{3.73\%} & \textbf{1.79\%} & \textbf{1.75\%} & \textbf{1.54\%}&\textbf{0.41\%}&\textbf{0.29\%} & 480 & \textbf{0.000\%}&\textbf{0.000\%} &\textbf{0.023\%}\\
        \hline
        FunSearch* & 4.48\%& 4.07\%& 3.02\% & 2.06\%&1.99\% &0.55\%&0.31\%& 464& 0.000\%&0.000\% &0.029\%\\
        \hline
        FunSearch  & 5.30\% & 4.19\%& 3.11\%& 2.47\%& -&0.68\%&0.32\%&\textbf{512}& - & -&-\\
        \hline
        EoH  & -& -& -& -&2.24\%&0.80\%&0.61\%& - & 0.000\%&0.000\% &0.025\%\\
        \hline
        
    \end{tabular}
    }
    \caption{Main experiment results on each task. The best result for each setting is in \textbf{bold}. Our method outperforms "FunSearch*", our reproduction of FunSearch on all problems, and is better than FunSearch on online bin packing as well as EoH on TSP.}
    \label{tab: main_result}
\end{table*}

\subsection{Main Results}
In Table \ref{tab: main_result}, we report the performance of the best heuristics acquired by each method. Our method significantly outperforms all baseline methods in all datasets of OBP. The fraction of excess bins cost by our methods is 9.36\% $\sim$ 41.73\% lower than ``FunSearch*" and 10.98\% $\sim$ 42.44\% lower compared with results reported in FunSearch on OR datasets. Despite the high performance of baseline methods on generated Weibull distribution instances, our method can still outperform baseline methods. For TSP, even though all methods are very close to the optimal solution, our method still performs better than other baseline methods, with the gap with the optimal route 20.69\% smaller than ``FunSearch*" and 8.00\% than EoH. Both result demonstrates the quality of heuristics acquired using our method, with non-trivial performance improvement in these tasks despite already high performing baselines.

Our method outperforms ``FunSearch*" in the cap sets problem, where we find a cap set that is greater than ``FunSearch*" by 16 for n=8. Although we are not able to surpass the performance reported in FunSearch \cite{romera2024mathematical}, we argue it's too hard to reproduce their results due to the extremely high time and computational cost of a complete cap set experiment, making it impossible for us to run as many times as FunSearch \footnote{Running a cap set experiment requires generating and evaluating 2.5 million programs, it takes more than 3 days on our GPU server. As stated in \cite{romera2024mathematical}: among 140 experiments they ran on cap set problem with n=8, less than 5\% yield cap set larger than 480. It is extremely computationally heavy to try to reproduce the result they reported.}. Yet, our method can find larger cap sets than ``FunSearch*". We believe it is sufficient to demonstrate the superiority of our method even on extremely difficult tasks against baseline methods.

\subsection{Discussion}
Since the performance of the best run might be influenced by randomness, we carry out some experiments to prove the performance gain is due to our method's efficacy in both exploitation and exploration.
We use the OR library of OBP as the target problem in this section.

\begin{figure}
    \centering
    \includegraphics[width=\linewidth]{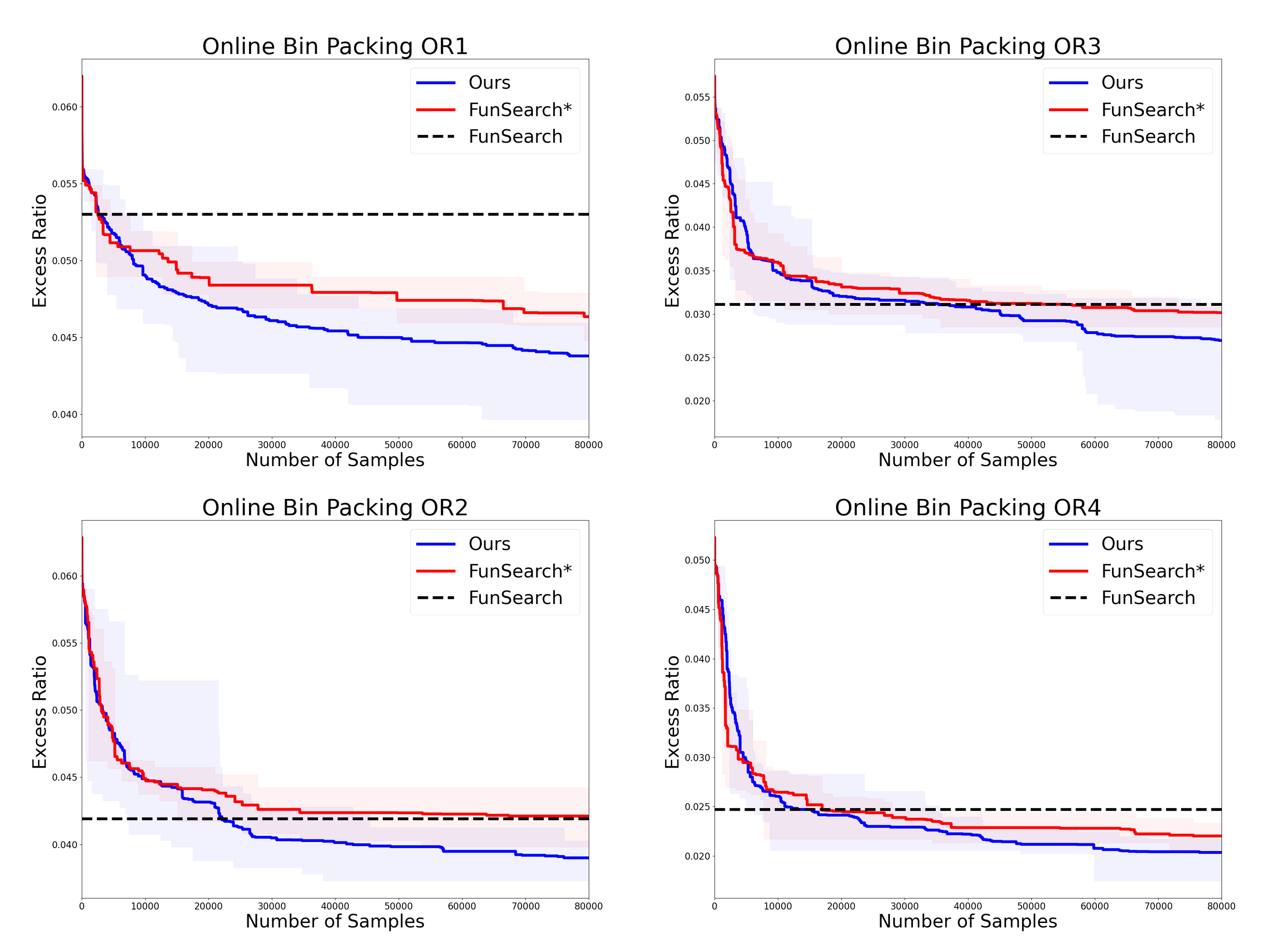}
    \caption{Performance progress on online bin packing. The solid line shows the average score among 10 experiments at each timestep. The shadow shows the range of best and worst experiments. FunSearch is shown in dash line since only a final score is available.}
    \label{fig: bp score}
\end{figure}
In Figure \ref{fig: bp score}, we show the performance progress of our method compared with ``FunSearch*", our replica of FunSearch, as only the final score is available for the original FunSearch. The solid lines represent the average progress of each method, with the shaded regions indicating the range from the best to the worst run. On average, our method (blue) outperforms both FunSearch (dashed black) and FunSearch* (red) at an early stage.

As shown in Figure \ref{fig: recentbest}, the ``Recent Best Score" of our method consistently surpasses that of ``FunSearch*". Our method demonstrates significant performance improvements even in later stages, whereas ``FunSearch*" encounters plateaus. This indicates that our method can steadily exploit the current state to achieve further gains, while FunSearch struggles to do so. We attribute this advantage to our priority criterion, which aligns more closely with the goal of exploitation.

In Figure \ref{fig: recentpoc}, we present the ``Recent Proportion of Change" for both ``FunSearch*" and our method. The new samples generated by our method consistently exhibit lower similarity to their parents compared to those of FunSearch, indicating that our method explores a broader region of the function space overall. Furthermore, our method gradually reduces exploration over time, allowing more opportunities for exploitation, consistent with our analysis in Section \ref{analysis}. In contrast, FunSearch demonstrates relatively low and indiscriminate exploration, which is less effective.

At the same time, the pace of improvement of the best sample's score in our method is higher than the baseline, with a relatively significant increase in the later stages. This suggests that our method can balance exploitation and exploration, which in all leads to stable performance improvements, and eventually outperforms baselines in the long term.

\subsection{Ablation Study}

\label{ablation}
\begin{table}[]
    \centering
    \resizebox{0.95\columnwidth}{!}{
    \begin{tabular}{|l|c|c|c|c|}
        \hline
         &Parent & UIIS& Best & $\mathrm{Avg_{std}}$ \\
        \hline
        \hline
        Ours& $\Tilde{Q}_p(\mathbb{C},t)$ & True & \textbf{1.79\%} & $\textbf{2.76\%}_{0.0016}$\\
        Parent Selection Only& $\Tilde{Q}_p(\mathbb{C},t)$& False& 2.65\%& $\textrm{2.89\%}_{0.0018}$ \\
        Quality Only& $Q_t(\mathbb{C})$& False&  2.74\%& $\textrm{2.98\%}_{0.0012}$\\
        \hline
        FunSearch* & $s(\mathbb{C})$& False& 3.02\% & $\textrm{3.07\%}_{0.0008}$ \\
        \hline
        \end{tabular}}
    \caption{Ablation of our method on online bin packing OR3. “Best” stands for the smallest excess rate acquired among 10 runs. ``Avg$_{\mathrm{std}}$" stands for the average score, with standard deviation shown as the suffix.}
    \label{tab: ablate_method}
\end{table}

We carried out an ablation study to provide a deeper understanding of QUBE. Experiments are carried on the OR3 dataset of OBP. Unless otherwise specified, all methods (variants) share the same implementation as Section \ref{implement detail}.
Several variants of our method experimented with are:

\noindent\textbf{Parent Selection Only}: ``Parent Selection Only" adopts the same parent selection as our method, with clusters with top-2 $\Tilde{Q}_t(\mathbb{C})$ are chosen for parents at each timestep. Its island reset strategy is the same as FunSearch.

\noindent\textbf{Quality Only}: ``Quality Only" selects clusters with top-2 $Q_t(\mathbb{C})$ for parents at each timestep. This measure of sample quality does not involve uncertainty. Its island reset strategy is the same as FunSearch.

We report the best as well as average excess rate (along with standard deviation) among 10 runs for each variant in Table \ref{tab: ablate_method}. 

The performance gap between ``FunSearch*" and "Quality Only" showcases the importance of using $Q_t(\mathbb{C})$ instead of $s(\mathbb{C})$ as the evaluation of the sample's quality. The reason behind this result is that $Q_t(\mathbb{C})$ is an unbiased estimate of the expected outcome with offspring samples in $\mathbb{C}$ serving as parents, while $s(\mathbb{C})$ is not despite being more intuitively straightforward. This leads to better exploitation of our method than FunSearch.

Comparing the results of ``Quality Only" and ``Parent Selection Only", we see further performance gains.  The integration of uncertainty into UIQ allows samples within rarely chosen clusters to be selected as parents. This allows our method to explore areas in the ``function spaces" that may evolve better samples despite not seeming promising at present.
Therefore, our method automatically balances between exploration and exploitation and eventually benefits the long-term performance.

Furthermore, our island reset procedure resets islands that are unlikely to evolve high-score programs, in contrast to FunSearch that reset islands that have relatively low score at present. Our method keeps islands with the potential of evolving better samples, while FunSearch is short-sighted. The performance difference between ``Ours" and ``Parent Selection Only" provides evidence of the rationality of our island reset procedure.


\subsection{Choice of LLMs}

\begin{table}[]
    \centering
    \resizebox{0.95\columnwidth}{!}{
    \begin{tabular}{|l|l|c|c|}
        \hline
         LLM& Method & Best Run & $\mathrm{Avg_{std}}$ \\
        \hline
        \hline
         \multirow{2}{*}{OpenCoder} & FunSearch* & 3.02\% & $\textrm{3.07\%}_{0.0008}$\\
         & Ours & \textbf{1.79\%} & $\textbf{2.76\%}_{0.0016}$\\
        \hline
         \multirow{2}{*}{Deepseek} & FunSearch* & 3.09\%& $\textrm{3.19\%}_{0.0011}$\\
         & Ours & 2.69\%& $\textrm{2.89\%}_{0.0017}$\\
        \hline
    \end{tabular}}
    \caption{Different LLM's result on online bin packing OR3. Our method steadily performs better than FunSearch, regardless of alternations in LLM.}
    \label{tab: ablate_llm}
\end{table}

To check if the performance gain from our method is invariant to unrelated conditions like LLM, we carry out experiment on OR3 dataset of OBP. Apart from OpenCoder-8b-Instruct \cite{huang2024opencoder} used in experiments before, we select another LLM with a smaller size and possibly lower code generation performance namely Deepseek-coder-6.7b \cite{Guo2024deepseekcoder}. We show results in Table \ref{tab: ablate_llm}. 

The result shows that our method always leads to better performance than FunSearch, even when a LLM with poor performance is used. which justifies it as model agnostic. Moreover, the result acquired from OpenCoder is always better than Deepseek-coder, which is a weaker LLM in comparison. Such results suggest that utilizing larger or better LLMs, even better results on hard problems like cap set may be possible.

\section{Conclusion}
In this paper, we studied FunSearch, a type of LLM+EA method that optimizes heuristics through evolution. We discovered that it has significant drawbacks: not doing well in either exploitation or exploration. Inspired by UCB, we propose our method QUBE that can address this issue. Experiment results demonstrate that our method steadily outperforms baseline methods, regardless of the task or unrelated conditions like specific LLM. We are optimistic that, boosted by our method, FunSearch can fully utilize LLM's potential and further be able to solve more complex problems in an even wider range of fields.

\section{Limitations}
Despite making non-trivial improvements on combinatorial optimization problems like online bin packing and TSP, our method fails to outperform heuristics searched by FunSearch \cite{romera2024mathematical} on the cap set problems. Although this may potentially diminish the superiority of our method on large-scale complex problems, we have made every effort to demonstrate the advantage of our method over ``FunSearch*" on the cap set problem under comparable settings. The performance of the best heuristics discovered is related to the choice of LLM, the number of samples generated and some random factors. Besides, to the best of our knowledge, no research work has ever surpassed or even tested the result of FunSearch \cite{romera2024mathematical} in the cap set problem due to its extremely high computation requirements. We see this as an opportunity to further extend the capability and efficiency of LLM+EA methods.

Moreover, our method as well as FunSearch, requires generating codes using LLMs and running these codes on some devices. This might be dangerous, since the code generated by LLM may be unpredictable and hard to explain. In our experiment, we observed codes generated by LLM trying to modify (write and read) local files. We tried our best to overcome this risk in our experiments by restricting permission to access local disk, running codes in safe namespaces, etc.
\bibliography{custom}

\newpage
\appendix

\section{More Experiment Details}
\subsection{Construction of Data}
\label{appn: more exp detail}
We list further details of our experiments here.

For OR datasets of online bin packing, we directly run our method and baseline methods on the test instances of each subset (OR1 $\sim$ OR4). The offline lower bound for each instance in these datasets is available, and the excess ratio for each subset is calculated directly using the sum of all used bins and the sum of all lower bounds of all instances. 

For Weibull datasets of online bin packing, we generate 5 test instances for each setting following settings in \cite{romera2024mathematical}, with 1k, 5k, 10k items each for Weibull1k, Weibull5k, Weibull10k respectively. Each bin's capacity is set to 100. The size of each item is sampled from Weibull(45, 3) distribution, clipped to 0$\sim$100, and finally rounded to an integer between 1 and 100. The offline lower bound for each instance in Weibull datasets is calculated following \cite{martello1990lower}.

The input for the cap set problem is simply the number of dimensions $n$. Since the cap set problem is already solved for $n\leq 6$, we experimented with $n=8$. Our method generates a heuristic within a guided greedy construction of cap set. Each heuristic can be evaluated through the size of the cap set found using itself.

The test instances for TSP are generated following the same setting as previous works \cite{kool2018attention,liu2024evolution}. For each setting (TSP20, TSP50, TSP100) 1000 test instances are generated, each with 20, 50, or 100 locations randomly initialized from $[0,1]^2$, respectively.

\subsection{Hyperparameter Setting}
Apart from implementation details mentioned in Section \ref{implement detail}, we list the hyperparameter settings in Table \ref{tab: fun_detail}. One hyperparameter, specifically $k$ used in Equation \ref{eq: uiq} for UIQ, is searched for the optimal value since it influences the overall performance significantly. We show the results in Appendix \ref{appn: hyperparam}. The values of other hyperparameters are either identical to FunSearch \cite{romera2024mathematical} or carefully chosen to ensure the results are suitable for our implementation and hardware while also comparable among baselines.

\begin{table*}[]
    \centering
    
    \begin{tabular}{|l|l|c c|c|c|}
        \hline 
         & Hyperparameter & \multicolumn{2}{|c|}{OBP} & Cap Set & TSP\\
         & & OR & Weibull & & \\
        \hline 
        \hline
        LLM Samplers & Number of samplers  & 16&16 & 16 & 16\\
         & LLM nucleus sampling $p$& 0.95 & 0.95 & 0.95 & 0.95 \\
         & LLM sampling temperature $t$ & 1.0 & 1.0 & 1.0 & 1.0 \\
         & Samples generated per prompt: $n_s$ & 4& 4& 4 & 1\\
         & Total number of samples & 80K& 80k& 2M & 2K\\
        \hline 
        Evaluators & Number of evaluators & 50& 50& 50 & 50\\
         & Timeout limit (in seconds) & 30& 60& 90 & 90\\
        \hline 
        DataBase & Number of islands: $n$ & 10& 10& 10 & 1\\
         & UIQ hyperparameter for uncertainty: $k$ \quad\quad& 0.0008& 0.0001& 32.0 & $10^{-5}$\\
         & Island reset interval: $T_{reset}$ & 32,768& 32,768& 262,144 & -\\
         & Temperature for choosing sample: $T_{prog}$ & 1.0& 1.0& 1.0 & 1.0\\
         \hline
    \end{tabular}
    \caption{Implementation details for our method as well as baseline methods. }
    \label{tab: fun_detail}
\end{table*}

\section{More Results for Figure 1}
\label{appn: fig1 more}
\begin{figure}
    \centering
    \includegraphics[width=\linewidth]{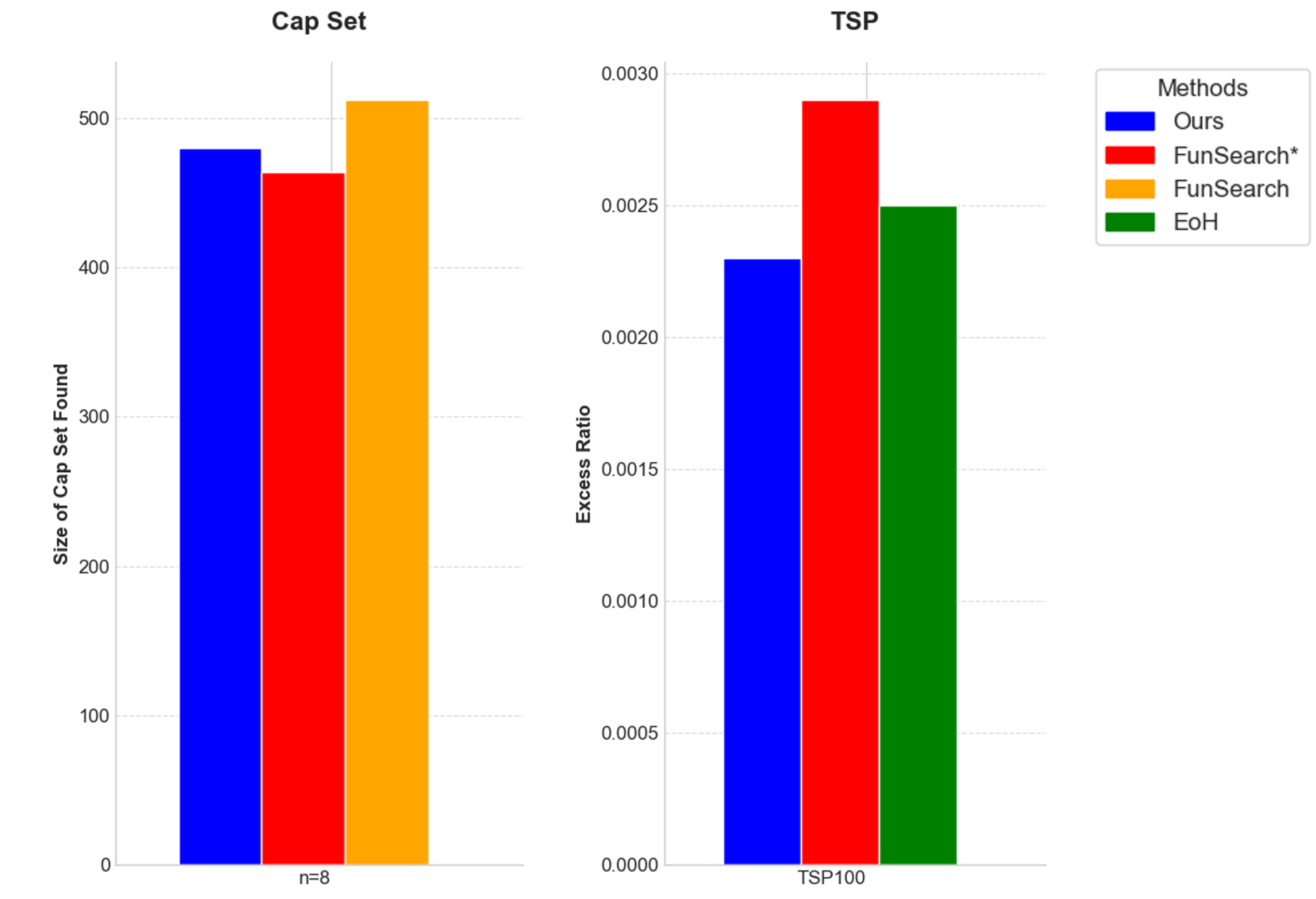}
    \caption{More experiment results on cap set n=8 and TSP100. For TSP a smaller excess ratio is better, while for cap set a larger found set size is better.Our method still shows superiority over baseline methods.}
    \label{fig: appn intro more}
\end{figure}

In Figure \ref{fig: figure1} of Section \ref{introduction}, we only show experiment results on online bin packing. We plot more experiment results in Figure \ref{fig: appn intro more}. Our method finds a larger cap set than ``FunSearch*" and outperforms all baseline methods on TSP100. Since the result on TSP20 and TSP50 is all 0 for all method, which is equal to the theoretical best, we are not showing them in plots.

\section{Hyperparameter Search Results}
\label{appn: hyperparam}

\begin{table}[!b]
    \centering
    \begin{tabular}{|c|c|c|}
    \hline
        $k$ &Best Run& Avg \\
    \hline
    \hline
    0.01 & 2.87\%& 2.97\%\\\hline
    0.008& 2.84\%& 3.05\%\\\hline
    0.004 & 2.97\%& 3.03\%\\\hline
    0.002 & 2.89\%& 3.12\%\\\hline
    0.001 & 2.74\%& 2.86\%\\\hline
    0.0008 &\textbf{2.59\%} &\textbf{2.79\%} \\\hline
    0.0004 & 2.72\%& 2.84\%\\\hline
    0.0002 & 2.68\%& 2.82\%\\\hline
    0.0001 & 2.70\%& 2.89\%\\\hline
    \end{tabular}
    \caption{Hyperparameter search result for $k$ on OR3 online bin packing. The optimal $k$ is 0.0008.}
    \label{tab: appn kp bp or}
\end{table}
The value for the hyperparameters used in our method, namely UIQ's hyperparameter $k$, is searched. To search the best value for $k$, we run experiments on ``UIQ-only" method as described in Section \ref{ablation}. Apart from the cap set problem, each setting is run 10 times to calculate the average performance.

For OR dataset of OBP, we investigated that the appropriate value for $k$ should be between 0.01 to 0.0001 so as to balance the quality term and uncertainty term well. Experiments are run on OR3 dataset. We provide experiment results for $k$ in Table \ref{tab: appn kp bp or}. 

For Weibull dataset of OBP, we investigated that the appropriate value for $k$ should be between 0.001 to 0.00001 so as to balance the quality term and uncertainty term well. Experiments are run on Weibull5k dataset. We provide experiment results for $k$ in Table \ref{tab: appn kp bp wb}. 

\begin{table}[!b]
    \centering
    \begin{tabular}{|c|c|c|}
    \hline
        $k$ &Best Run& Avg \\
    \hline
    \hline
    0.001 & 1.73\%& 1.86\%\\\hline
    0.0008& 1.65\%& 1.90\%\\\hline
    0.0004 & 1.67\%& 1.83\%\\\hline
    0.0002 & 1.62\%& 1.75\%\\\hline
    0.0001 & \textbf{1.54\%}& \textbf{1.72\%}\\\hline
    0.00008 & 1.59\% &1.79\% \\\hline
    0.00004 & 1.64\%& 1.82\%\\\hline
    0.00002 & 1.60\%& 1.78\%\\\hline
    0.00001 & 1.70\%& 1.88\%\\\hline
    \end{tabular}
    \caption{Hyperparameter search result for $k$ on Weibull5k online bin packing. The optimal $k$ is 0.0001.}
    \label{tab: appn kp bp wb}
\end{table}

Similarly, for cap set problem, we experimented $kr$ within the range of 16 to 64. Since it cost heavily to run cap set experiments, we only run 5 runs for each setting and show the results in Table \ref{tab: appn kr cs}. 

\begin{table}[]
    \centering
    \begin{tabular}{|c|c|c|}
    \hline
        $k$ &Best Run& Avg \\
    \hline
    16 & 464& 452.8\\\hline
    32 & \textbf{464} &\textbf{464} \\\hline
    48 & 464& 451.2\\\hline
    64 & 448& 448\\\hline
    \end{tabular}
    \caption{Hyperparameter search result for $k$ on cap set n=8. We use UIQ-only for experiment. The optimal value is 32.}
    \label{tab: appn kp cs}
\end{table}

\section{Code Specification for Each Task}
\label{appn: specification}

In this section, we show the code specifications for each task. The function decorated with ``@evolution" is evolved in experiments and the score of each function can be acquired by running the function decorated with ``@run" on each test instance.

For online bin packing, the code specification is shown in Table \ref{tab: appn bp spe}. For the cap set problem the code specification is shown in Table \ref{tab: appn capset spe}. For TSP, the code specification is shown in Table \ref{tab: appn tsp spe}.

\section{Best Heuristics Discovered}
We show the best heuristics discovered by our method for each task here. The whole part of the function LLM samplers outputs are shown without any modification, which is why some part of the answers might sound nonsense.

For online bin packing OR1 the best heuristic discovered is shown in Table \ref{tab: appn or1 best}. For OR2, the best heuristic is shown in Table \ref{tab: appn or2 best}. For OR3, the best heuristic is shown in Table \ref{tab: appn or3 best}. For OR4, the best heuristic is shown in Table \ref{tab: appn or4 best}.

For cap set n=8, our best heuristic finds a cap set of 480 vectors. The corresponding heuristic is shown in Table \ref{tab: appn cn8 best}.

\begin{table*}[]
    \centering
    \begin{tabular}{c}
         \begin{python}
import os 
import numpy as np

class BinPackProblem:
  def __init__(self, id, capacity, n_items, best_answer, items):
    self.id = id
    self.capacity = capacity
    self.n_items = n_items
    self.best_answer = best_answer
    self.items = np.array(items)
    assert len(items) == n_items
    bins = [capacity] * n_items
    self.bins = np.array(bins)
  
def get_valid_bin_indices(item, bins: np.ndarray) -> np.ndarray:
  return np.nonzero((bins - item) >= 0)[0]
  
def online_binpack(items: tuple[float, ...], bins: np.ndarray) -> tuple[list[list[float, ...], ...], np.ndarray]:
  packing = [[] for _ in bins]
  for item in items:
    valid_bin_indices = get_valid_bin_indices(item, bins)
    priorities = priority(item, bins[valid_bin_indices])
    best_bin = valid_bin_indices[np.argmax(priorities)]
    bins[best_bin] -= item
    packing[best_bin].append(item)
  packing = [bin_items for bin_items in packing if bin_items]
  return packing, bins

@run
def evaluate_binpack(problem):
  items = problem.items
  bins = problem.bins
  best_answer = problem.best_answer
  capacity = problem.capacity
  _, bins_packed = online_binpack(items, bins)
  solved_answer = (bins_packed != capacity).sum()
  cnt = best_answer - solved_answer
  ratio = cnt / best_answer
  return ratio

@evolution
def priority(item: float, bins: np.ndarray) -> np.ndarray:
  # Returns the priority with which we want to add 'item' to the bins
  return 0.0
         \end{python}
    \end{tabular}
    \caption{Code specification for online bin packing.}
    \label{tab: appn bp spe}
\end{table*}

\begin{table*}[]
    \centering
    \begin{tabular}{c}
         \begin{python}
"""Finds large cap sets."""
import itertools
import numpy as np

def solve(n: int) -> np.ndarray:
  """Returns a large cap set in `n` dimensions."""
  all_vectors = np.array(list(itertools.product((0, 1, 2), repeat=n)), dtype=np.int32)
  # Powers in decreasing order for compatibility with `itertools.product`, so
  # that the relationship `i = all_vectors[i] @ powers` holds for all `i`.
  powers = 3 ** np.arange(n - 1, -1, -1)
  # Precompute all priorities.
  priorities = np.array([priority(tuple(vector), n) for vector in all_vectors])
  # Build `capset` greedily, using priorities for prioritization.
  capset = np.empty(shape=(0, n), dtype=np.int32)
  while np.any(priorities != -np.inf):
    # Add a vector with maximum priority to `capset`, and set priorities of
    # invalidated vectors to `-inf`, so that they never get selected.
    max_index = np.argmax(priorities)
    vector = all_vectors[None, max_index]  # [1, n]
    blocking = np.einsum('cn,n->c', (- capset - vector) 
    priorities[blocking] = -np.inf
    priorities[max_index] = -np.inf
    capset = np.concatenate([capset, vector], axis=0)

  return capset

@run
def evaluate(n: int) -> int:
  """Returns the size of an `n`-dimensional cap set."""
  capset = solve(n)
  return len(capset)

@evolution
def priority(element: tuple[int, ...], n: int) -> float:
  """Returns the priority with which we want to add `element` to the cap set."""
  return 0.0
         \end{python}
    \end{tabular}
    \caption{Code specification for cap set problem.}
    \label{tab: appn capset spe}
\end{table*}

\begin{table*}[]
    \centering
    \begin{tabular}{c}
         \begin{python}
import numpy as np
import random
import math

def euclidean_distance(city1, city2):
    return math.sqrt((city1[0] - city2[0])**2 + (city1[1] - city2[1])**2)

def calculate_total_distance(route, distance_matrix):
    return sum(distance_matrix[route[i]][route[i+1]] for i in range(len(route)-1)) + distance_matrix[route[-1]][route[0]]

def two_opt(route, distance_matrix):
    best_route = route.copy()
    improved = True
    while improved:
        improved = False
        for i in range(1, len(route)-2):
            for j in range(i+1, len(route)):
                if j-i == 1: continue
                new_route = route[:i] + route[i:j][::-1] + route[j:]
                if calculate_total_distance(new_route, distance_matrix) < calculate_total_distance(best_route, distance_matrix):
                    best_route = new_route
                    improved = True
        route = best_route
    return best_route

@run
def guided_local_search(cities, max_iterations=100, alpha=0.1):
    num_cities = len(cities)
    distance_matrix = np.zeros((num_cities, num_cities))
    for i in range(num_cities):
        for j in range(i+1, num_cities):
            distance_matrix[i][j] = distance_matrix[j][i] = euclidean_distance(cities[i], cities[j])
    init_distance_matrix=copy.deepcopy(distance_matrix)
    # Initialize route
    route = list(range(num_cities))
    best_route=route
    # Initialize penalties
    penalties = np.zeros((num_cities, num_cities))
    for iteration in range(max_iterations):
        # Local search with 2-opt
        route = two_opt(route, distance_matrix)
        # Update route
        if calculate_total_distance(route, init_distance_matrix) < calculate_total_distance(best_route,init_distance_matrix):
            best_route=route
        # Evolve distance_matrix
        distance_matrix=distance_matrix+update_dist(distance_matrix,best_route)
    return best_route, calculate_total_distance(best_route, init_distance_matrix)

@evolution
def update_dist(distance_matrix, current_route):
    ''' calculates an update to current distance matrix. '''
    return np.zeros_like(distance_matrix)
         \end{python}
    \end{tabular}
    \caption{Code specification for TSP.}
    \label{tab: appn tsp spe}
\end{table*}

\begin{table*}[]
    \centering
    \begin{tabular}{c|c}
         \begin{python}
def priority(item: float, bins: np.ndarray) -> np.ndarray:
  penalty_factor_v3 = 0.7

  D_item_val, C_int_fit, B_valid_region, a_of_K2_val = 4.5, 3.5, 2.6, 4.7

  item_weight = item / 4650

  scores = np.zeros(len(bins))

  K_values = np.array([0.28, 0.31, 0.35])

  B_values = np.array([0.15, 0.3, 0.25])

  b_weights = np.array([2750/4650, 2950/4650, 3050/4650, 3150/4650])

  for index, bin_num in enumerate(bins):
    quantity_1D = index * bin_num
    calc_2D_quantity = bin_num * bin_num

    if index <= 3400:
      b_weight = b_weights[0]
    elif index<=3800:
      b_weight = b_weights[1]
    else:
      b_weight = b_weights[3]

    P_item = (index * b_weight) * (quantity_1D / calc_2D_quantity)

    # Further improvements here.

    improved_P_item = P_item * (index**52) * (item_weight**67) * (index**2.5) * (item_weight**4.0) * (index**3.4) * (item_weight**3.2) * (index**3.0) * (item_weight**3.3)

    valid_region = abs(quantity_1D / calc_2D_quantity - 1)

    if index <= 3000:
      K = (K_values[0] * penalty_factor_v3) + ((1 - penalty_factor_v3) * K_values[1])
    elif index<=3800:
      K = K_values[1]
    else:
      K = K_values[2]

    if index <= 3500:
      B_val = (B_values[0] * penalty_factor_v3) + ((1 - penalty_factor_v3) * B_values[1])
    elif index<=3800:
      B_val = B_values[1]
    else:
      B_val = B_values[2]

    intersection_fit = ((index * item_weight / (abs(bin_num - item)))**42) * K * 2400000

    improved_D_item_val = D_item_val * ((bins[index]/item) ** 2.8) * (1.0 + index / 95000)
    improved_C_int_fit = C_int_fit * (95 / (index+6))
    improved_B_valid_region = B_val + (1-B_val) * (valid_region**2.5)
    improved_a_of_K2_val = a_of_K2_val / (1 + index / 95000)

    P_final = improved_D_item_val * ((improved_P_item + C_int_fit * intersection_fit) / (improved_B_valid_region * (improved_a_of_K2_val + valid_region)))

    scores[index] = P_final

  return scores
         \end{python}
    \end{tabular}
    \caption{The best heuristic searched by our method for OR1 online bin packing.}
    \label{tab: appn or1 best}
\end{table*}

\begin{table*}[]
    \centering
    \begin{tabular}{c|c}
         \begin{python}
def priority(item: float, bins: np.ndarray) -> np.ndarray:            
  bins_difference = np.abs(bins - item)

  low_threshold, high_threshold = 8, 23
  diff_mid = (high_threshold + low_threshold) / 2

  p_vect4 = np.where(bins_difference <= low_threshold, bins_difference * (-1) * 22,
                  np.where(bins_difference <= diff_mid, bins_difference * (-1) * 34,
                  np.where(bins_difference <= high_threshold, bins_difference * (-1) * 46, bins_difference * (-1) * 2)))

  p_vect4[np.abs(bins_difference) <= high_threshold / 2] += 35
  p_vect4[np.abs(bins_difference) <= diff_mid] += 50
  p_vect4[np.abs(bins_difference) <= low_threshold + high_threshold / 2] += 64

  for i, val in enumerate(bins_difference):
    if val <= 25:
        bins_difference[i] = bins_difference[i] * (i + 1) * 72
    else:
        break

  if np.any(np.abs(np.where(bins_difference <= 25, bins_difference * (-1) * 100, bins_difference * (-1) * 13)) <= 150):
    p_vect4[np.abs(np.where(bins_difference <= 25, bins_difference * (-1) * 95, bins_difference * (-1) * 13)) <= 150] += 42

    best_global = sorted(p_vect4)
    best_three_values = best_global[0:3]
    worst_bin_index = np.where(p_vect4 == max(best_three_values))[0][0]

    if worst_bin_index < len(p_vect4):
      p_vect4[worst_bin_index] = min(p_vect4) * 0.98

  return p_vect4 
         \end{python}
    \end{tabular}
    \caption{The best heuristic searched by our method for OR2 online bin packing.}
    \label{tab: appn or2 best}
\end{table*}

\begin{table*}[]
    \centering
    \begin{tabular}{c|c}
         \begin{python}
def priority(item: float, bins: np.ndarray) -> np.ndarray:
  probabilities = np.zeros(len(bins), dtype=float)

  for i in range(len(bins)):
    current_bin_space = bins[i]

    if item <= current_bin_space:
      remainingSpaceFactor = current_bin_space / (current_bin_space + item)
      enhanced_load_factor = item/current_bin_space

      # Improved estimation formula: f(x) = a * x ** p * exp(x)

      """
      Non-uniform impact approach based on the load intensity:
      Enhance the evaluated importance of loading by approaching loader-bins outcomes.
      """
      additional_impact_factor = 0.00

      if enhanced_load_factor < 0.95:
        modified_priority = (0.99 * ((remainingSpaceFactor / (1 - enhanced_load_factor)) - 2.55 + additional_impact_factor) * 1500 - 95 / (remainingSpaceFactor ** 1.25)) * (130 + 0.0095 * i) * np.exp(-i * 0.022)
      elif enhanced_load_factor < 0.99:
        modified_priority = (1.00 * ((remainingSpaceFactor / (1 - enhanced_load_factor)) - 2.45 + additional_impact_factor) * 1600 - 45 / (remainingSpaceFactor ** 1.30)) * (140 + 0.0105 * i) * np.exp(-i * 0.022)
      else:
        modified_priority = (1.01 * ((remainingSpaceFactor / (1 - enhanced_load_factor)) - 2.35 + additional_impact_factor) * 1700 - 35 / (remainingSpaceFactor ** 1.35)) * (160 + 0.0115 * i) * np.exp(-i * 0.023)

      # Added/displaced non-uniform interpolated/smooth kernel-duty system aspects

      modified_priority -= 500 + 70 * np.cos(enhanced_load_factor + 0.07) + 600 * np.tanh(2.84 * (enhanced_load_factor - 0.93)) + 80 * np.cos(2 * i / len(bins)) + 880 * np.sin(2 * i / len(bins))

      # Adjust differently for injected non-trivial items using maximum performance complexity system

      modified_priority -= 35 * (1-enhanced_load_factor) ** 0.98

      # Insert updated, optimized weights for different scenarios

      probabilities[i] = modified_priority

  return probabilities 
         \end{python}
    \end{tabular}
    \caption{The best heuristic searched by our method for OR3 online bin packing.}
    \label{tab: appn or3 best}
\end{table*}

\begin{table*}[]
    \centering
    \begin{tabular}{c|c}
         \begin{python}
def priority(item: float, bins: np.ndarray) -> np.ndarray:            
  def improved_prior_func(_value):
    if _value < item / 9:
      if bins.size > 700:
        return 260**(35 * item / 350 - 2.5 * _value)
      elif bins.size > 350:
        return 140**(30 * item / 350 - 1 * _value)
      else:
        return 140**(50 * item / 350 - 2.5 * _value)   # Colocalization

    elif _value < item / 5:
      if bins.size > 700:
        return 180**(35 * item / 350 - 1 * _value)
      elif bins.size > 350:
        return 110**(40 * item / 350 - 0.5 * _value)  #Quorum sensing
      else:
        return 140**(40 * item / 350 - 0.6 * _value)   # Quorum sound BiellLIF

    elif _value < item:
      if bins.size > 700:
        return 95 * item /(145 + item)
      elif bins.size > 350:
        return 80 * item /(125 + item)
      else:
        return 80 * item /(130 + item)    #Rotulina colleague asymmetrically restructuring translators replication achieved in cell-process

    else:
      if bins.size > 700:
        return 105 * item /(130 + item)
      elif bins.size > 350:
        return 95 * item /(120 + item)
      else:
        return 95 * item /(110 + item)       #Biulation sncRNA oscillations

  return np.vectorize(improved_prior_func)(bins - item) 
         \end{python}
    \end{tabular}
    \caption{The best heuristic searched by our method for OR4 online bin packing.}
    \label{tab: appn or4 best}
\end{table*}
\begin{table*}[]
    \centering
    \begin{tabular}{c|c}
         \begin{python}
def solve(n: int) -> np.ndarray:
  score = np.sum(element) * 220.00 * 3.0
  zeros = [idx for idx, val in enumerate(element) if val == 0]
  # If there are at least two zeros.
  if len(zeros) >= 2:
    score = np.abs(np.sum(zeros)) * 230.00 * 2400.0
  # If there are at least three zeros.
  if len(zeros) >= 3:
    d = np.array(zeros)[1:] - np.array(zeros)[:-1]
    d_sorted = np.sort(d)
    r = d_sorted[-1]
    if r 
      score = np.abs(zeros[0] - zeros[1]) * 250.00 * 3400.0
  # If there are at least four zeros.
  if len(zeros) >= 4:
    score = np.sum(element) * 260.50 * 35.0
  # If there are more than three zeros and less than six zeros.
  if len(zeros) > 3 and len(zeros) < 6:
    score += 35000.0 * np.sum(zeros)
  # If there are more than five zeros and less than nine zeros.
  if len(zeros) > 5 and len(zeros) < 9:
    score += 36000.0 * np.sum(element)
  # If there are six or more zeros.
  if len(zeros) >= 6:
    score *= np.sum(np.array(element))
  # Add some score based on the minimum and maximum elements.
  score += np.sum(element) * np.min(np.array(element[:2])) * np.max(np.array(element)) * 100.00
  # If there is one zero, multiply the score by 120.
  if len(zeros) == 1:
    score *= 120.0
  # Subtract some value based on the sum of the elements.
  score -= np.sum(element) * np.sum(element[:2]) / 4.5
  # If there are no zeros, multiply the score by 115.
  if len(zeros) == 0:
    score *= 1.15
  # Multiply the score by 40.
  score *= 40.00
  # If there are seven or more zeros, add some value to the score.
  if len(zeros) >= 7:
    score += np.sum(element) * 250.00 * 120.0
    score *= 1.85
  if len(zeros) > 9 and len(zeros) < 12:
    score += np.sum(element) * 260.50 * 90.0
  # If there are twelve or more zeros, add some value to the score.
  if len(zeros) >= 12:
    score += np.sum(element) * 280.50 * 140.0
  if len(zeros) > 14:
    score *= np.sum(zeros)
  # Multiply the score by the maximum element plus 40.
  score *= np.max(np.array(element)) + 40.00
  if np.sum(element) <= 12:
    score *= 1.75
  # If there are five or fewer zeros, multiply the score by 27.
  if len(zeros) <= 5:
    score *= 27.0
  # Add 12000 to the score.
  score += 12000.0
  # If there are ten or fewer zeros, add 20000 to the score.
  if len(zeros) <= 10:
    score += 20000.0
  # If there are fifteen or fewer zeros, add 30000 to the score.
  if len(zeros) <= 15:
    score += 30000.0
  # Further improved version of `priority_v2`.
  score *= 1.75
  # Final improvement of the score.
  score *= 1.45
  return score
         \end{python}
    \end{tabular}
    \caption{The heuristic searched by our method that leads to a cap set of size 480 on n=8.}
    \label{tab: appn cn8 best}
\end{table*}
\section{LLM Prompts}
\label{appn: prompt}
We write task-specific natural instructions for LLM samplers in MarkDown style, since the LLM we choose is capable of understanding and generating in MarkDown style. In all prompts shown below, ``\{Parent1\}" and ``\{Parent2\}" are replaced with two parents selected at each time step.

For online bin packing, the prompt we use is shown in Table \ref{tab: appn prompt bp}.
For cap set problem, the prompt we use is shown in Table \ref{tab: appn prompt cs}.
For TSP, the prompt we use is shown in Table \ref{tab: appn prompt tsp}.

\begin{table*}[]
    \centering
    \begin{tabular}{p{0.98\textwidth}}
\toprule
Online 1D bin packing problem is a combinatorial optimization problems. The goal of online bin packing is to assign each of a series of items into the smallest number of fixed-sized bins. Generally, heuristics are used to solve online bin packing efficiently. Priority function is defined in heuristic to help rank and search for best candidates. \\
You are given two priority functions "priority\_v0" and "priority\_v1", then you are asked to complete the following priority function "priority\_v2" such that it is an improved version of "priority\_v1". This priority function will be used in heuristic to ranks the priority of bins given incoming item. \\

Here are the requirements:\\
1. Just complete the "priority\_v2" function and do note answer anything else.\\
2. Do not use "print" function in your answer.\\
\\
``` python\\
\# Finds good assignment for online 1d bin packing.\\
import numpy as np\\
\\
def priority\_v0(item: float, bins: np.ndarray) -> np.ndarray:\\
\quad\quad  """ Returns the priority with which we want to add 'item' to the bins """\\
\{Parent1\}\\
\\
def priority\_v1(item: float, bins: np.ndarray) -> np.ndarray:\\
\quad\quad  """ Improved version of priority\_v0 """\\
\{Parent2\}\\
\\
def priority\_v2(item: float, bins: np.ndarray) -> np.ndarray:\\
\quad\quad  """ Improved version of priority\_v1 """\\

\bottomrule
    \end{tabular}
    \caption{Prompt Template for online bin packing}
    \label{tab: appn prompt bp}
\end{table*}

\begin{table*}[]
    \centering
    \begin{tabular}{p{0.98\textwidth}}
\toprule
The cap set problem calculates the largest possible set of vectors in \$\\mathbb\{Z\}\^n\_3\$ (known as a cap set) such that no three vectors sum to zero. Geometrically, no three points of a cap set lie on a line.\\
Generally, heuristics can be used to solve cap set problem. Priority function for solving the cap set problem ranks the priority with which we want to add a vector into the cap set.\\
Given two priority functions "priority\_v0" and "priority\_v1" where "priority\_v1" is an improved version of "priority\_v0", your task is to complete the following function priority\_v2 such that it is an improved version of priority\_v1. Just complete the code and do not answer anything else. Do not use any `print` function in your answer.\\
\\
Here are the requiremnets:\\
1. Just complete the "priority\_v2" function and do note answer anything else.\\
2. Do not use "print" function in your answer.\\
\\
``` python\\
\# Find large cap sets\\
import numpy as np\\
import itertools\\
def priority\_v0(n: int) -> np.ndarray:\\
\quad\quad """ Returns a large cap set in 'n' dimensions."""\\
\{Parent1\}\\
\\
def priority\_v1(n: int) -> np.ndarray:\\
\quad\quad  """ Improved version of priority\_v0 """\\
\{Parent2\}\\
\\
def priority\_v2(n: int) -> np.ndarray:\\
\quad\quad  """ Improved version of priority\_v1 """\\

\bottomrule
    \end{tabular}
    \caption{Prompt Template for cap set problem}
    \label{tab: appn prompt cs}
\end{table*}

\begin{table*}[]
    \centering
    \begin{tabular}{p{0.98\textwidth}}
\toprule
TSP problem finds shortest paths that travels all places and return to the starting point. Guided local search can be used to iteratively update solution to TSP problems. A function updates the distance matrix according to current shortest paths, such that further local search on the updated distance matrix may lead to better answer. \\
You are given two update functions "update\_dist\_v0" and "update\_dist\_v1", then you are asked to complete the following priority function "update\_dist\_v2" such that it is an improved version of "update\_dist\_v1". This priority function will be used in heuristic to ranks the priority of bins given incoming item. \\

Here are the requirements:\\
1. Just complete the "update\_dist\_v2" function and do note answer anything else.\\
2. Do not use "print" function in your answer.\\
\\
``` python\\
import numpy as np\\
import random\\
import math\\
import copy\\
\\
def update\_dist\_v0(distance\_matrix ,current\_route):\\
\quad\quad  """ Updates the distance matrix according to current best route searched"""\\
\{Parent1\}\\
\\
def update\_dist\_v1(distance\_matrix ,current\_route):\\
\quad\quad  """ Improved version of update\_dist\_v0 """\\
\{Parent2\}\\
\\
def update\_dist\_v2(distance\_matrix ,current\_route):\\
\quad\quad  """ Improved version of update\_dist\_v1 """\\

\bottomrule
    \end{tabular}
    \caption{Prompt Template for TSP.}
    \label{tab: appn prompt tsp}
\end{table*}

\end{document}